\documentclass[11pt,a4paper]{article}
\usepackage[hyperref]{emnlp2018}
\usepackage{times}
\usepackage{latexsym}
\usepackage{amsmath}
\usepackage{booktabs}
\usepackage{listings}
\usepackage{graphicx}
\usepackage{tabularx}

\usepackage{url}

\usepackage{color}
 
\definecolor{codegreen}{rgb}{0,0.6,0}
\definecolor{codegray}{rgb}{0.5,0.5,0.5}
\definecolor{codepurple}{rgb}{0.58,0,0.82}
\definecolor{backcolour}{rgb}{0.94,0.94,0.94}

\lstdefinestyle{mystyle}{
    backgroundcolor=\color{backcolour},   
    commentstyle=\color{codegreen},
	otherkeywords={self},             
	keywordstyle=\ttfamily\color{blue!90!black},	
	keywords=[2]{True,False},
	keywords=[3]{ttk},
    basicstyle=\footnotesize,
    breakatwhitespace=false,         
    captionpos=b,                    
    keepspaces=true,                 
    numbers=none,                    
    numbersep=3pt,                  
    showspaces=false,                
    showstringspaces=false,
    showtabs=false,                  
    tabsize=2,
    frame=single
}
 
\aclfinalcopy 

\newcommand\confname{EMNLP 2018}

\title{Instructions for \confname{} Proceedings}

\title{INFODENS: An Open-source Framework for Learning Text Representations}

\author{Ahmad Taie$^{1}$ \and  Raphael Rubino$^{1,2}$ \and Josef van Genabith$^{1,2}$ \\
  Saarland University, Germany$^{1}$ \\
German Research Center for Artificial Intelligence (DFKI), Germany$^{2}$\\
  {\tt ahmad.g.taie@gmail.com} \\ {\tt \{raphael.rubino, josef.van\_genabith\}@dfki.de}}

\date{}

\begin{document}
\maketitle
\begin{abstract}
The advent of representation learning methods enabled large performance gains on various language tasks, alleviating the need for manual feature engineering. While engineered representations are usually based on some linguistic understanding and are therefore more interpretable, learned representations are harder to interpret. Empirically studying the complementarity of both approaches can provide more linguistic insights that would help reach a better compromise between interpretability and performance.
We present INFODENS, a framework for studying learned and engineered representations of text in the context of text classification tasks. It is designed to simplify the tasks of feature engineering as well as provide the groundwork for extracting learned features and combining both approaches. INFODENS is flexible, extensible, with a short learning curve, and is easy to integrate with many of the available and widely used natural language processing tools.
\end{abstract}

\section{Introduction}
\label{section:introduction}
Linear classifiers in combination with the right features achieve good performance on text classification tasks \cite{WangManning}. Those hand-crafted features provide baselines for evaluating deep learning methods and are sometimes difficult to beat \cite{zhang2015character,ConneauSBL16}. In some cases, hand-crafted features can even be combined with learned features to improve performance on a given task \cite{joinemDasha,sennrich-haddow:2016:WMT} highlighting some complementarity in the information captured by each approach. Conducting empirical experiments to study such complementarity would be beneficial, and the reasons are threefold: Firstly, this enables us to compare the performance of both hand crafted and learned representations and make design decisions regarding the trade-offs between speed and accuracy on a specific dataset. Secondly, it helps in investigating where the performance gaps are and whether these methods can complement each other and how they can be combined to improve performance. Finally, it allows us to derive new linguistic hypotheses as in many cases, deep learning methods are great engineering tools but they operate as black box methods and it is difficult to extract from them linguistic insights.

In this paper we present INFODENS ~\footnote{Code and documentation available at the project's repository: \texttt{github.com/ahmad-taie/infodens}} a framework aimed at studying hand-crafted and learned representations. We first explain how INFODENS can be used to simplify the tasks of feature engineering, feature learning, and evaluation. We then validate the framework on sentiment analysis and topic classification tasks and showcase that in many cases, hand-crafted features can be complementary to learned representations.

\section{Framework Design and Architecture}
\label{section:toolkit_design_and_architecture}
The framework is designed in a modular and developer-friendly manner to encourage changes and extensions. The source code is accompanied by a user and a developer guide, and we give a brief overview of the architecture in this section, summarized in Figure~\ref{figure:architecture}.
The framework consists of the following frozen and hot spots:

\begin{figure}[!h]
	\begin{center}
		\includegraphics[scale=0.4]{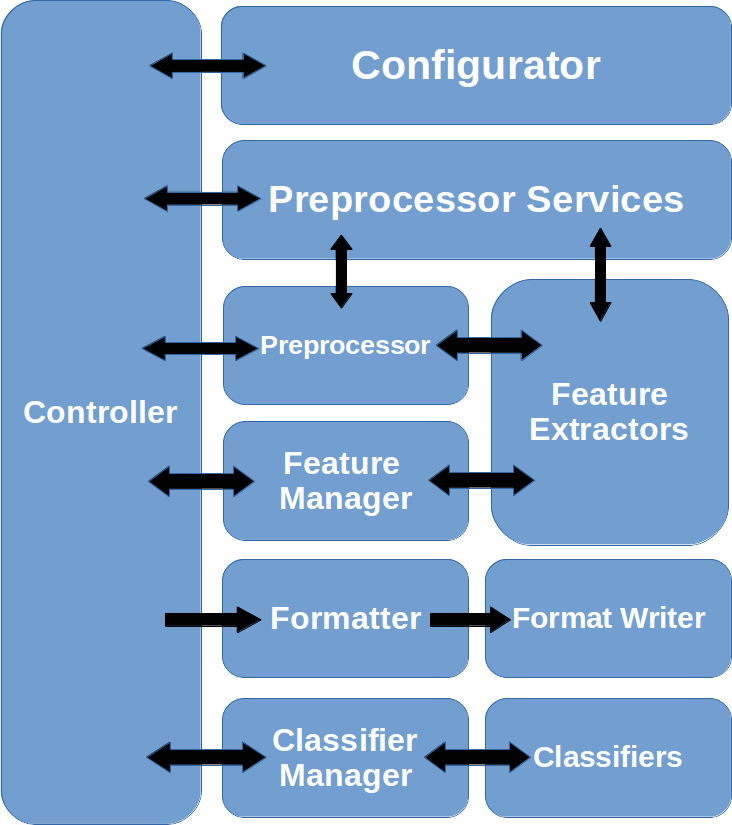} 
		\caption{Overview of the framework's architecture.}
		\label{figure:architecture}
	\end{center}
\end{figure}

\subsection{Frozen spots}

These are the modules of the framework that need not be changed for extending the functionality in typical use cases.

\begin{description}
\item [Controller] is the callable module and centerpiece of the framework. It instantiates the other modules, calls their APIs, and handles the communication between them.
\item [Preprocessor] provides the APIs for accessing the input text, preprocessed versions of it, and external resources. It also handles the building of language models and the unsupervised learning of word-embeddings. 
\item [Feature manager] dynamically detects the available feature extractors and manages the multi-threaded feature extraction process. It handles merging the extracted and given feature matrices and generating feature descriptors.
\item [Formatter] is the module that exports the extracted features in a chosen format. This can also be extended with other existing or custom formats via the Format writer. ~\footnote{Currently, the extracted features can be exported in CSV, and in the input formats for the \textsc{weka}~\cite{frank2016weka} and \textsc{libsvm}~\cite{chang2011libsvm} toolkits.} 
\item [Classifier manager] manages the training and evaluation of the different classifiers or regressors. Like the feature manager, it also detects the classifiers dynamically at run time.

\end{description}

\subsection{Hot spots}

These are the modules which developers can modify and extend with their code to add new functionality.

\begin{description}
\item [Preprocessor services] is used to integrate different NLP tools (taggers, tokenizers.. etc) without changing the Preprocessor APIs. It can also be called to do on-the-fly preprocessing of feature-specific input files.
\item [Configurator] handles the definition and extraction of configuration parameters from configuration files.

\item [Feature extractors] extract and return vector representations of text, whether learned or engineered. Researchers can write their own feature extractor methods which are detected dynamically at run-time and called by the feature manager.

\item [Classifiers] are trained on the extracted features to build a model that is then used to evaluate the features. Their design is inspired by the \textsc{scikit-learn} \cite{sklearn_api} approach. Similar to the feature extractors, they are detected dynamically by the classifier manager.

\item [Format writer] implements the feature output formats. It can be extended to support other formats by adding new methods to the class. 

\end{description}

\section{Usage}
\label{section:framework_usage}

The framework can be used as a standalone toolkit without any modifications given the implemented features and classifiers. For example, it can be used to extract features for usage with other machine learning tools, or to evaluate given features with the existing classifiers or regressors. Extending the framework with new feature extractors or classifiers is as simple as a drag and drop placement of the new code files into the \texttt{feature\_extractor} and \texttt{classifer} directories respectively. The framework will then detect the new extensions dynamically at runtime. In this section we explore how each use case is handled.

\subsection{Feature Extraction and Evaluation}

The framework is run by invoking the Python script \texttt{infodens.py} with an INI configuration file consisting of five sections specifying the input files, the output parameters, the general settings, the requested features and their arguments, and finally, the classifiers.
Figure~\ref{figure:config} shows an example of a configuration file. All the parameters are described in the README file on the repository.

\begin{figure}[!h]

\begin{lstlisting}
[Input]
train file : data/sent2.train
train classes:  data/sent2_classes.train
test file : data/sent2.test
test classes:  data/sent2_classes.test

train feats: feats_trainFeats.libsvm
test feats: feats_testFeats.libsvm

training corpus: data/literature.orig.en
language : eng

[Output]
classifier report: report.txt
output features: feats2 libsvm

[Settings]
kenlm : kenlm/bin
threads : 4

[Classifiers]
Decision_tree
SVC_linear: -rank 5
Keras_MLP: -hidden_layers 10,5 -epochs 2

[Features]
1
4: -ngram 1 -cutoff 2
\end{lstlisting}
    \caption{Configuration file with all the sections and some parameters}
    \label{figure:config}
\end{figure}

\subsection{Feature Development}

Since a main use case for the framework is extracting engineered and learned features, it was designed such that developing a new feature extractor would require minimal effort. Figure~\ref{figure:featEx} demonstrates a simple feature extractor that retrieves the sentence length. More complicated features and learned features are provided in the repository which can be used as a guide for developers. Documentation for adding classifiers and format writers is described in the Wiki of the repository but is left out of this paper due to the limited space.

\begin{figure*}[!ht]
	\centering{
\begin{lstlisting}[language=Python]
from infodens.feature_extractor.feature_extractor import featid, Feature_extractor
from scipy import sparse

class Surface_features(Feature_extractor):
	
    def getSentLen(self, sentences):
        sentLen = []
        for sent in sentences:
            sentLen.append(len(sent))
        return sparse.lil_matrix(sentLen).transpose()

    @featid(10)
    def sentenceLength(self, argString, preprocessReq=False):
        '''Returns length of every sentence. '''

        if preprocessReq:
            # Request all preprocessing functions to be prepared
            self.preprocessor.gettokenizeSents()
            self.testPreprocessor.gettokenizeSents()
            return 1

        sentLen = self.getSentLen(self.preprocessor.gettokenizeSents())
        testSentLen = self.getSentLen(self.testPreprocessor.gettokenizeSents())

        return sentLen, testSentLen, "Sentence Length"
\end{lstlisting}

    \caption{Example feature extractor. A feature class that inherits from \texttt{Feature\_extractor} is defined which can then contain multiple feature extractors. A feature extractor is a method within that class with a decorator \texttt{@featid} that assigns it a unique numeric ID. The extractor is given 2 parameters, a string for arguments passed to it in the configuration file, and a flag for a preprocessing run. The extractor must first check if this call to it is a preprocessor call (\texttt{preprocessReq} is \texttt{True}), which is used to gather the different preprocessing requests from all extractors, so as not to repeat the requests. It then returns. When the preprocessing request is false, the extractor must then proceed to call the preprocessor APIs and retrieve the required train and test data to fill in and return 2 SciPy sparse matrices with sizes $(n1,x)$ and $(n2,x)$ where $x$ is the length of the feature vector, and $n1$ and $n2$ are the number of sentences in the train and test sets respectively. The extractor also returns as a third parameter a string describing the feature vector.}
    \label{figure:featEx}
    }
\end{figure*}

\begin{table*}[t]
\centering
\small
\begin{tabular}{@{\hspace{3pt}}l@{\hspace{3pt}}cccccccc}
\toprule
Dataset && AG & DBP & Yelp P. & Yelp F. & Yah. A. & Amz. F. & Amz. P. \\
\midrule
Number of training samples && 120k & 560k & 560k & 650k & 1.4M & 3M & 3.6M \\
\midrule
Number of classes && 4 & 14 & 2 & 5 & 10 & 5 & 2 \\
\midrule
ngrams~\cite{zhang2015character}       && 92.0 & 98.6 & 95.6 & 56.3 & 68.5 & 54.3 & 92.0 \\
char-CNN~\cite{zhang2015text}          && 87.2 & 98.3 & 94.7 & 62.0 & 71.2 & 59.5 & 94.5 \\
char-CRNN~\cite{xiao2016efficient}     && 91.4 & 98.6 & 94.5 & 61.8 & 71.7 & 59.2 & 94.1 \\
VDCNN~\cite{ConneauSBL16}               && 91.3 & 98.7 & 95.7 & 64.7 & 73.4 & 63.0 & 95.7 \\
fastText~\cite{joulin2016bag}     && 92.5 & 98.6 & 95.7 & 63.9 & 72.3 & 60.2 & 94.6 \\
\midrule
\midrule
\textsc{infodens} features: \\
N-grams (1-5) 		&& 92.4  & 98.8 & 93.1 & 58.4 & 71.9 & 57.5 & 94.4 \\
POS N-grams (1-5) 		&& 68.0  & 82.7 & 78.6 & 42.7 & 36.2 & 41.6 & 76.8 \\
\midrule
Hand-crafted representations: \\
Surface and lexical		&& 26.9 & 13.3 & 50.0 & 20.5 & 10.2 & 20.0 & 50.0 \\
Language model and surprisal && 27.6 & 8.7 & 56.1& 23.9 & 12.0 & 22.7 & 53.7 \\
POS Language model and surprisal && 27.2 & 12.8 & 50.1& 20.0 & 12.6 & 21.9 & 52.4 \\
N-gram frequency quantiles		&& 35.9 & 30.2 & 55.0 & 23.7 & 19.0 & 23.8 & 53.1 \\
Hashed n-grams (1-5)		&& 92.0  & 98.6 & 95.0 & 57.1 & 69.6 & 56.3 & 93.7 \\

Hashed POS n-grams (1-5) 		&& 66.2  & 81.8 & 80.2 & 42.8 & 37.1 & 32.0 & 58.3 \\
\midrule
Learned representations: \\
fastText		&& 91.8  & 98.1 & 95.6 & 55.8 & 69.4 & 49.6 & 94.4 \\
Average sentence embedding && 90.0 & 92.9 & 52.1 & 28.2 & 40.0 & 34.0 & 54.7 \\
\midrule
Hand-crafted (SVM)		&& 90.9  & 98.5 & 94.6 & 56.7 & 68.4 & 54.5 & 93.5 \\
Hand-crafted (MLP 100h)		&& 90.8 & 98.5 & 94.7 & 59.3 & 64.5 & 58.4 & 94.0 \\
Hand-crafted (MLP 100,50h)		&& 90.7  & 98.5 & 94.8 & 59.9 & 67.5 & 58.6 & 94.2 \\
\midrule
Learned	representation (SVM) && 92.0 & 98.0 & 93.7 & 46.7 & 67.8 & 48.1 & 93.5 \\
Learned	representation (MLP 100h) && 92.0 & 98.5 & 95.5 & 59.6 & 63.3 & 54.0 & 94.4 \\
Learned	representation (MLP 100,50h) && 92.0 & 98.4 & 95.5 & 62.4 & 66.5 & 55.8 & 94.4 \\
\midrule
All (SVM)		&& 92.3 & 98.8 & 95.3 & 56.3 & 68.3 & 53.1 & 94.4 \\
All (MLP 100h)		&& 91.8  & 98.7 & 95.1 & 60.2 & 66.5 & 58.6 & 94.7 \\
All (MLP 100,50h)		&& 91.9  & 98.7 & 95.2 & 59.7 & 68.7 & 58.7 & 94.7 \\

\bottomrule
\end{tabular}
\caption{Accuracy [\%] results on the test sets.}\label{table:all_results}
\end{table*}
\section{Evaluation and Results}
\label{section:toolkit_evaluation}
In this section, we evaluate the performance of the framework used out of the box. We first detail the datasets used, then the set of hand-crafted and learned representations, along with the classifiers, all of which are available as part of the released code.

\subsection{Datasets and External Resources}

We use the the datasets provided by Zhang et al. ~\shortcite{zhang2015character}, three of which are topic classification datasets: AG's news, DBpedia, and Yahoo! Answers, and four are for sentiment analysis: Yelp review polarity, Yelp review full, Amazon review polarity, and Amazon review full. We exclude the Sougu News dataset, which is a transliterated Chinese text, as we only utilize English language models and word embeddings for the purposes of this demonstration. The results gathered by \cite{joulin2016bag}, comparing different convolutional models and the fastText approach, are used as baselines.
External resources required to extract $n$-gram probabilities and word embeddings, namely a $5$-gram modified Kneser-Ney smoothed language model~\cite{kneser1995improved} and a set of skip-gram based word embeddings with $256$ dimensions~\cite{mikolov2013distributed}, are trained on a subset of the \textit{News Shuffle} corpus containing approx. $200$M sentences and $7.3M$ unique tokens~\cite{bojar-EtAl:2017:WMT1}.

\subsection{Hand-crafted Features}
We extract $5$ \textbf{Surface and Lexical features}, namely sequence length in number of tokens, average word length, type-token ratio, and lexical to tokens ratio (ratio of adjectives, verbs, nouns, and adverbs to tokens). \textbf{Bag of $n$-grams} features are extracted on the word and POS level. We use frequency cut-offs of $3,5,5,10,10$ for $n$-grams from $1$ to $5$ respectively for the smaller datasets and ten times higher for the Yahoo! and Amazon datasets. For \textbf{POS $n$-grams} we use cut-offs $10$ for unigrams and $20$ for bigrams and higher. For the Yahoo! and Amazon datasets we use cut-offs of $10,20,60,80,100$. The $n$-grams features are then also extracted using the hashing trick with the same cut-offs to reduce the final feature vector size when combined with other features. scikit-learn's \cite{scikit-learn} \texttt{FeatureHasher} is used with output vectors sizes of $5,7,7,10,15$ $\times$ $10^4$ for ngrams from $1-5$ respectively and $0.5,25,70,100,150$ $\times$ $10^2$ are used for POS ngrams. We extract lexical and POS level \textbf{Language model features} based on external language models, namely sentence log probabilities, perplexities, and surprisal in units of bits. Building the language model and extracting the features is done by providing the path to the compiled binaries for \textsc{kenlm}~\cite{Heafield11kenlm:faster}. Finally we extract \textbf{N-gram Frequency Quantile Distribution} features with the same cut-offs as in the bag of ngrams features, with $4$ quantiles and an OOV quantile. NLTK \cite{NLTK} is used for tokenization and POS tagging.
\subsection{Learned Features}

We extracted two features that use a learned representation: Firstly, we get a \textbf{sentence embedding} feature that is built by averaging the word embeddings of an input sentence. Secondly, we extract a \textbf{fastText representation} using the fastText library  with the same parameters as reported in Joulin et al. \shortcite{joulin2016bag}.

\subsection{Classifiers}

The linear SVC from scikit-learn \cite{scikit-learn} which is based on LIBLINEAR \cite{REF08a} is trained as a baseline for evaluating each feature type as well as the concatenated features. A grid search for $C$ is performed with $10$ values in the log scale ranging from $2^{-5}$ to $2^5$. Performance is then also compared to feeding the concatenated features into a feed-forward neural network. We report the results on two settings, a network with a single fully-connected hidden layer of size $100$ and another network with two fully-connected hidden layers of sizes $100$ and $50$ respectively. Both networks use a softmax output layer. The implementation is done using Keras \cite{chollet2015keras} with the TensorFlow \cite{TensorFlow} backend. The two smaller datasets and the Amazon datasets are trained for $2$ epochs and the remaining datasets are trained for $5$ epochs. We use with the Adam optimizer with a learning rate of $0.001$, and dropout with rate $0.3$. A single NVIDIA Titan X GPU is used for all experiments and time per epoch ranges from a few seconds for a small number of features on the smallest datasets to $5$ hours on the full feature set on the largest datasets. These settings were not chosen to optimize accuracy, but only for the purpose of evaluating the framework due to the large number of experiments presented. Users are encouraged to experiment with different hyper-parameters values and network sizes, as well as modify the code to build more sophisticated neural network models. Experimenting with the other classifiers available in the framework, such as logistic regression, can provide additional insightful comparisons.

\section{Results and Discussion}
We present the results in~Table~\ref{table:all_results}. In Zhang et al. ~\shortcite{zhang2015character} it was noted that the performance of ngram features degrades for larger datasets. However, we have seen in our baseline experiments that this effect can be reduced by using suitable frequency cut-offs. We have also seen that in many cases, the ngram features can solely outperform the neural approaches. For the two smaller datasets, linear classifiers tend to perform better, while for the larger datasets performance increases with increasing the non-linear layers even for hand-crafted representations. Combining hand-crafted and learned features is often beneficial, but not always, especially with the linear classifier.
What is clear is that different datasets benefit from different representations and model parameters and it is difficult to find a representation that consistently performs well across all datasets. This necessitates repeated experimentation to understand which approaches and parameters would provide more consistent improvements.

\section{Related Work}
\label{section:related_work}

While there exist toolkits such as \textsc{FEXTOR} \cite{Fextor}, \textsc{EDISON} \cite{Edison}, Learning Based Java \cite{RizzoloRo10}, and NLP frameworks such as GATE \cite{Cunningham2011a} that facilitate feature extraction, INFODENS differs in that it integrates feature learning in the extraction pipeline along with customizable feature evaluation.
Additionally, a main design goal of INFODENS is to require little to no programming experience to be used as a standalone toolkit, and minimal programming effort to develop new features and classifiers. This is accomplished as the framework is developed fully in Python, taking advantage of the plethora of libraries available for deep learning and natural language processing. And due to the interpreted nature of Python, extensions to the library require no recompilation and, by design, are discovered dynamically at runtime.

\section{Conclusions and Future work}
\label{section:conclusion}
We presented INFODENS, a framework aimed at learning text representations and showed how combining hand-crafted and learned representations can be beneficial. The framework provides flexible usage and extension scenarios enabling rapid evaluation of different text representations on different tasks. We aim to integrate more learned representations of text, namely convolutional features, and additionally, the next iteration of the framework will focus on allowing features to be combined differently, for example to be fed into different neural network layers, such as to an embedding or a convolutional layer instead of vanilla fully connected layers. Finally, a module to visualize the learned feature weights will be developed in order to understand which combination of features lead to a better classification decision.
\section*{Acknowledgments}
This  work  is  funded  by  the  German  Research Foundation (Deutsche Forschungsgemeinschaft) under grant SFB1102: Information Density and Linguistic Encoding.

\bibliography{infodens}
\bibliographystyle{acl_natbib_nourl.bst}

\end{document}